\documentclass[conference]{IEEEtran}
\IEEEoverridecommandlockouts
\usepackage{amsmath,amsfonts}
\usepackage{amssymb}
\usepackage{algorithmic}
\usepackage{algorithm}
\usepackage{array}
\usepackage[caption=false,font=normalsize,labelfont=sf,textfont=sf]{subfig}
\usepackage{textcomp}
\usepackage{stfloats}
\usepackage{url}
\usepackage{booktabs}
\usepackage{verbatim}
\usepackage{graphicx}
\usepackage{cite}
\usepackage[colorlinks, citecolor=blue]{hyperref}
\usepackage{multirow}
\usepackage{bbm}
\usepackage{bm}
\usepackage{bbding}
\usepackage{soul}
\usepackage{color}
\usepackage{xcolor}
\usepackage{float}
\usepackage{cleveref}
\newcommand{\std}[1]{\fontsize{5}{7}\selectfont {#1}}
\newcommand{\email}[1]{\href{mailto: #1}{\textcolor{black}{#1}}}
\def\BibTeX{{\rm B\kern-.05em{\sc i\kern-.025em b}\kern-.08em
    T\kern-.1667em\lower.7ex\hbox{E}\kern-.125emX}}

\begin{document}

\title{Representation Calibration and Uncertainty Guidance for Class-Incremental Learning based on Vision Language Model}

\author{Jiantao~Tan$^{\dagger}$,~Peixian~Ma$^{\dagger}$,~Tong~Yu,~Wentao Zhang,~Ruixuan~Wang

\thanks{Jiantao Tan, Tong Yu and Wentao Zhang are with the School of Computer Science and Engineering, Sun Yat-sen Univerisity, Gaungzhou 510275, China, and also with the Guangdong Province Key Laboratory of Machine Intelligence and Advanced Computing, Ministry of Education, Guangzhou 510275, China. (email: \{tanjt7, yutong3, zhangwt65\}@mail2.sysu.edu.cn).}

\thanks{Peixian Ma is with the Hong Kong University of Science and Technology (Guangzhou).~\email{pma929@connect.hkust-gz.edu.cn}.}

\thanks{Ruixuan Wang is with the School of Computer Science and Engineering, Sun Yat-sen Univerisity, Gaungzhou 510275, China, also with Peng Cheng Laboratory, Shenzhen 518066, China, and also with the Key Laboratory of Machine Intelligence and Advanced Computing, Ministry of Education, Guangzhou 510275,China. (e-mail: \email{wangruix5@mail.sysu.edu.cn}).}

\thanks{$^{\dagger}$~These authors contributed equally to this work.}

\thanks{Corresponding author: Ruixuan Wang.}

}


\maketitle

\begin{abstract}
Class-incremental learning requires a learning system to continually learn knowledge of new classes and meanwhile try to preserve previously learned knowledge of old classes. As  current state-of-the-art methods based on Vision-Language Models (VLMs) still suffer from the issue of differentiating classes across learning tasks. Here a novel VLM-based continual learning framework for image classification is proposed. In this framework, task-specific adapters are added to the pre-trained and frozen image encoder to learn new knowledge, and a novel cross-task representation calibration strategy based on a mixture of light-weight projectors is used to help better separate all learned classes in a unified feature space, alleviating class confusion across tasks. In addition, a novel inference strategy guided by prediction uncertainty is developed to more accurately select the most appropriate image feature for class prediction. Extensive experiments on multiple datasets under various settings demonstrate the superior performance of our method compared to existing ones.
\end{abstract}


\section{Introduction}

The rapid development of deep learning has led to impressive breakthroughs in various domains~\cite{ViT, CLIP, sqlr1}. However, unlike humans who can continually learn new knowledge from their experiences and adapt to new information over time, current deep learning systems typically rely on static datasets, and would catastrophically forget previously learned knowledge when they are learning new data~\cite{catastrophic-forgetting}.

Existing studies have been performed to empower deep learning models with the continual learning (CL) ability. Early CL studies often assume the models are initially trained from scratch and then continually updated with a sequence of CL tasks~\cite{TPAMI_survey}. 
Considering the strong generalization ability of pre-trained models,  researchers started to build CL systems on pre-trained vision or vision-language models (VLMs)~\cite{IJCAI_survey}. 
In particular, the integration of pre-trained textual information from VLM as supervision input significantly enhances the generalization capabilities of trained image features, increasing resilience to catastrophic forgetting.
Several CL methods based on pre-trained VLM (\textit{e.g.}, CLIP~\cite{CLIP}) have shown superior performance~\cite{PROOF, Lang-CL}. 
In VLM-based CL methods, while efficient fine-tuning of VLMs with task-shared light modules has shown promising performance~\cite{AttriCLIP, PR-CLIP}, they have been outperformed by the recently proposed task-wise fine-tuning approach~\cite{Lang-CL, yujun}, where each CL task is associated with a task-specific learnable light module. The main challenge in the task-wise fine-tuning approach is correct selection or appropriate fusion of task-specific modules or features.

\begin{figure}[t]
  \centering
  \includegraphics[width=0.47\columnwidth]{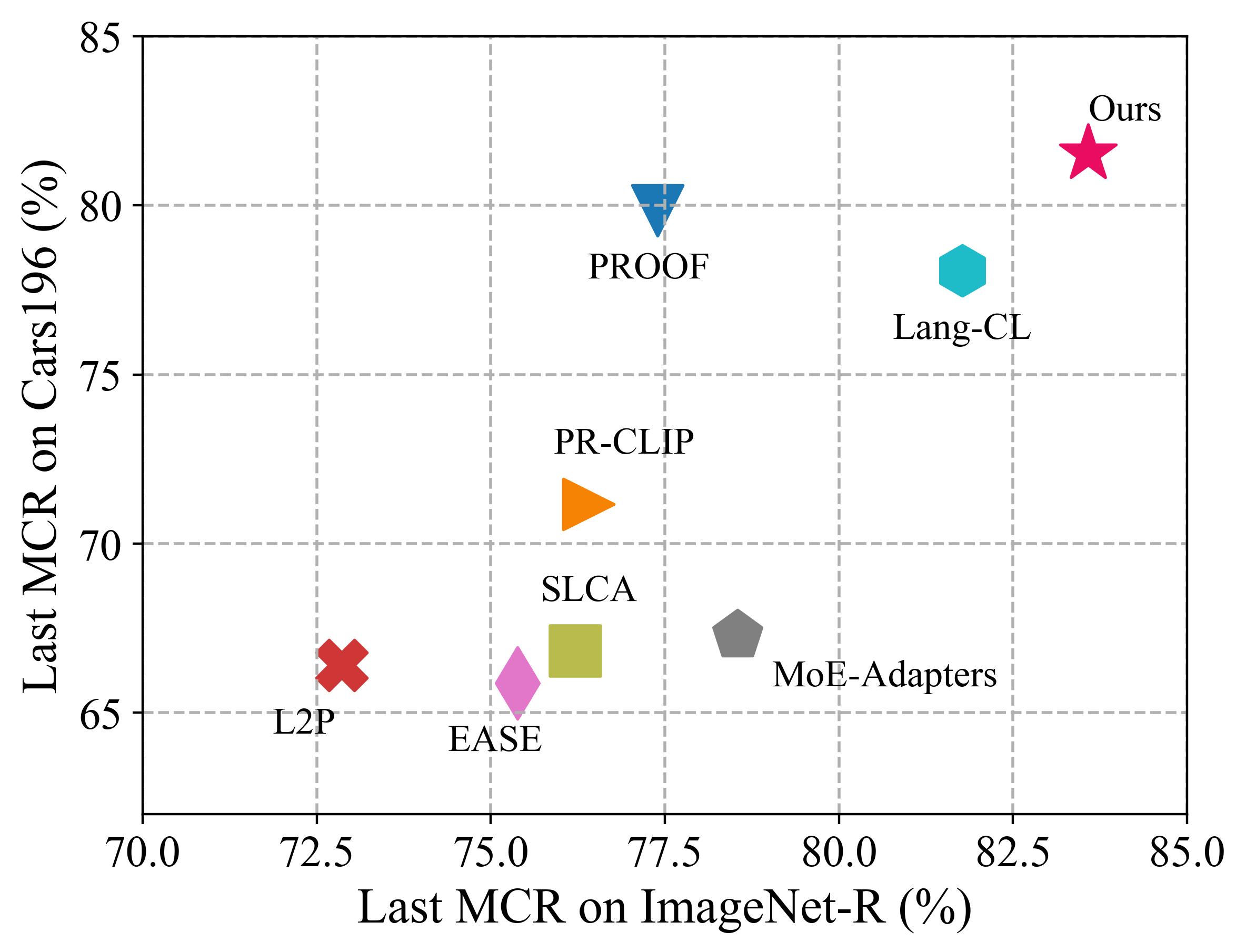}
  \includegraphics[width=0.47\columnwidth]{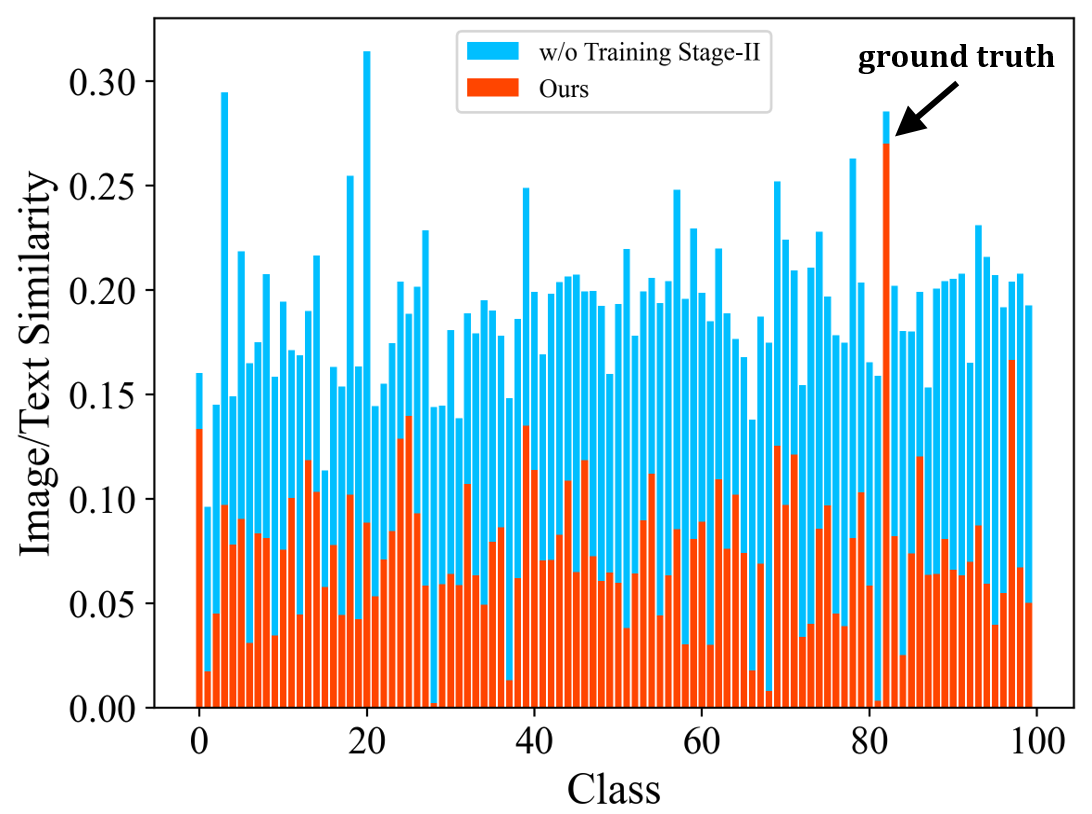}
  \caption{Left: Performance comparison of different methods on Cars196 and ImageNet-R in 10 task setting. Right: The image-text similarities of a class's image across all classes on CIFAR100 after incremental learning.}
  \label{fig:intro}
\end{figure}

\IEEEpubidadjcol

To solve the above challenge, a novel VLM-based framework for continual learning of image classes is proposed in this paper. The framework contains two novel designs. First, after learning a new task, a cross-task feature calibration strategy is introduced to help align independent task-specific feature spaces into a unified task-shared feature space, enhancing class separation across tasks. This is realized by training a novel mapping module called Mixture of Projectors (MoP) on top of the task-specific adapted VLM's image encoder. Second, a novel prediction uncertainty-guided inference strategy is utilized to select the most appropriate calibrated feature in task-shared space. This strategy is designed based on the prediction uncertainty that, if the task-specific image feature is extracted from the adapted image encoder of the task that test image belongs to, the prediction uncertainty of its corresponding calibrated feature transformed by MoP will be lower. Extensive empirical evaluations on multiple image classification benchmarks with various imaging domains and settings consistently support that the proposed CL framework is superior to existing approaches. The main contributions of this study are summarized below.

\begin{itemize}
    \item A novel VLM-based continual learning framework for image classification, where the cross-task feature calibration strategy can help unify task-specific feature spaces into a task-shared feature space for better class differentiations across tasks; 
    \item A novel inference strategy guided by model prediction uncertainty to help select the appropriate image feature for classification during inference; 
    \item Extensive evaluations on multiple benchmarks, with SOTA performance achieved from the proposed method.
\end{itemize}

\section{Related Work}

\noindent \textbf{Class-Incremental Learning:} Class-Incremental Learning (CIL) aims to enable learning systems to incrementally acquire knowledge from data of new classes over time. Approaches to CIL can be broadly categorized into four groups. Parameter regularization-based approaches~\cite{EWC} alleviate catastrophic forgetting by regularizing model parameters. Knowledge distillation-based approaches~\cite{LwF, WA, UCIR} use knowledge distillation technique to constrain the model’s outputs, helping the current model to mimic behavior of the previous model during updating. Replay-based approaches~\cite{iCarL,DarkER, BiC} usually maintain a memory buffer to store a small subset of old data for experience replay. Different from above approaches, structure-based approaches~\cite{DER, Lang-CL, EASE} usually add new modules to help learn knowledge of new classes, avoiding parameter overwriting and interference across tasks. 
Since this approach eliminates the need to address forgetting (stability), it primarily focuses on learning new data (plasticity) and selecting or merging parameters during inference.


\noindent \textbf{CIL on Vision-Language Models:} Vision-Language Models (VLMs) can leverage knowledge from text modality to support CIL of image classes, and therefore recent studies have started to focus on CIL with VLMs (especially with CLIP). These studies can be broadly divided into two types. One type involves fine-tuning the pre-trained image encoder~\cite{MoE-adapters, EASE,SSIAT}, usually adopting PEFT techniques. 
These methods still face the challenges of parameter overwriting or prompt selection. The other type involves freezing pre-trained CLIP image encoder and using outputs of the pre-trained image encoder for further learning. These methods typically enhance the image representations with trainable modules following the image encoder~\cite{PROOF, RAPF, CBM, tan2025augmenting}. But such frozen image features constrain learning performance on downstream CIL tasks especially when there is a large domain gap between downstream and pre-training data. This work addresses above issues by fine-tuning the encoder with task-specific parameters and proposes a better inference strategy.

\section{Methodology}
This study focuses on class-incremental learning (CIL). In CIL, a model continually learns new knowledge from a sequence of $T$ tasks, and each task contains a unique set of new classes. When the model is trained with the data of the $t$-th task ($t\leq T$), it often assumes training data of all previously learned (old) classes from task 1 to task $t-1$ are largely or totally unavailable. Here the challenging exemplar-free setting is adopted, \textit{i.e.}, no data of old classes are available when the model learns new knowledge from the current $t$-th task.

\begin{figure*}[!t]
  \centering
  \includegraphics[width=\linewidth]{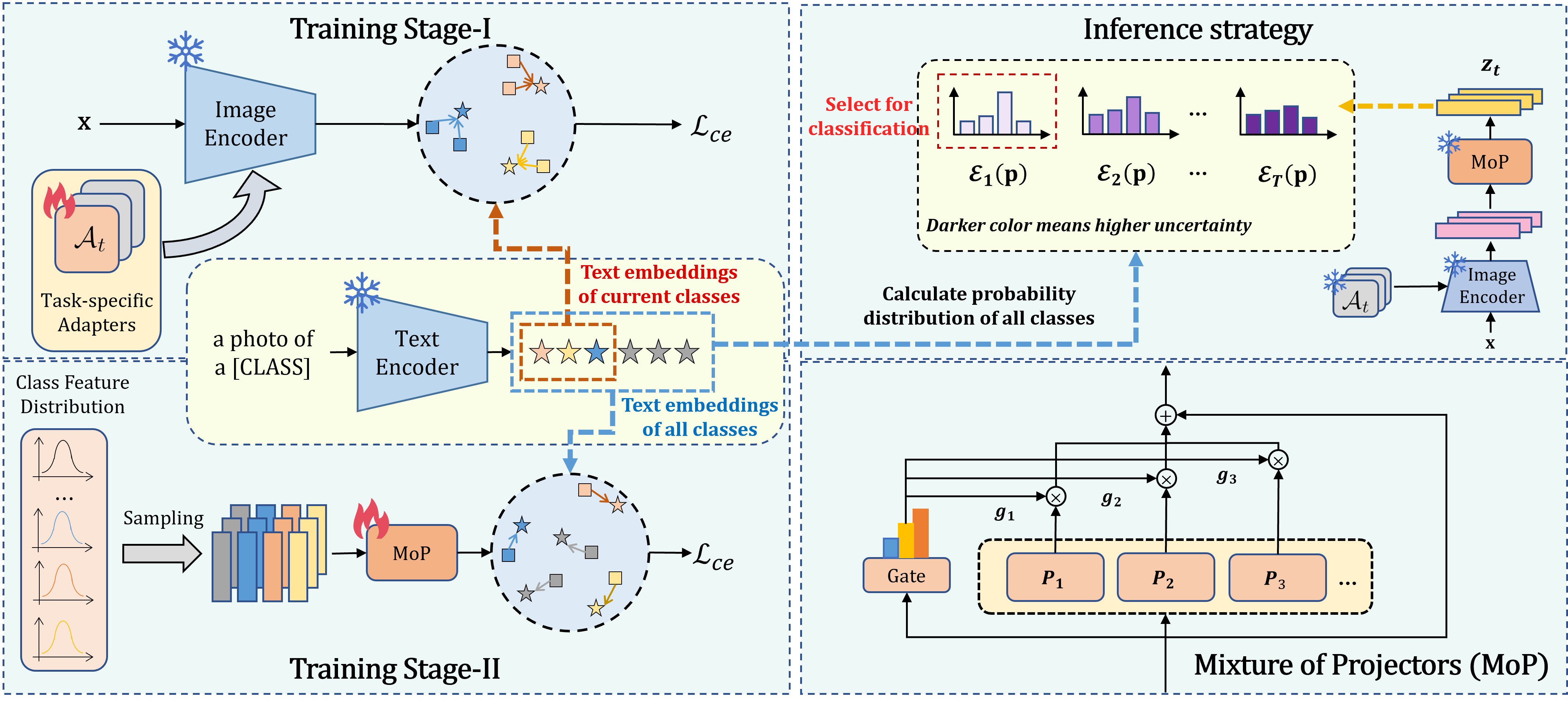}
  \caption{The proposed framework for exemplar-free class-incremental learning. For each new task, task-specific adapters are first trained (Training Stage-I), and then task-shared Mixture of Projectors (MoP) is optimized to help calibrate image embeddings from different task-specific adapted image encoders. A gating network in MoP adaptively adjusts image embeddings from each task-specific image encoder to help separate classes in the task-shared feature space. During inference, a novel feature selection strategy based on prediction uncertainty is proposed for more accurate class prediction.}
  \label{fig:method}
  \label{network}
\end{figure*}

\subsection{Two-stage Model Training}

\noindent \textbf{Task-specific adapters for CIL (Training Stage-I)}: 
When the model learns knowledge of $c_t$ new classes from the training set of the current $t$-th task, a light-weight adapter is added to each Transformer block of the CLIP's image encoder. The parameters of the original CLIP's image encoder are frozen and only the newly added task-specific adapters (denoted by $\mathcal{A}_t$) are optimized using the traditional cross-entropy loss $\mathcal{L}_{ce}$, where the $c_t$-dimensional probability output of the classifier is based on the image-text similarity. After learning all $T$ task, $T$ groups of adapters are obtained. During inference, since it is unknown which task the test image belongs to, we need to design an inference approach. The most straightforward approach \cite{Lang-CL} to solve this problem is to feed the test image through the image encoder of each task to obtain $T$ feature representations, compute their similarity scores with the corresponding class text features, and ultimately select the class corresponding to the highest similarity score as the prediction result. 

However, this inference strategy may cause a potential issue. For the adapted image encoder $f_i$ with the optimized task-specific adapters $\mathcal{A}_i$ in general can well classify the images from $i$-th task based on the image embeddings. But since no training images of other tasks are involved while training $\mathcal{A}_i$, the outputs of the image encoder $f_i$ for an image of task $j \neq i$ may be unexpectedly aligned well with the text embeddings of some classes from the $i$-th task. Such image-text misalignment could appear across any two tasks (As shown in the right of \Cref{fig:intro}, The blue bars show that in such inference approach, the similarities are generally high and difficult to distinguish), This would cause incorrect class prediction, since in such inference approach, one test image should passthrough the adapted encoder of task that it not belongs to.

\noindent \textbf{Cross-Task Representation Calibration (Training Stage-II)}: 
In order to alleviate such cross-task image-text misalignment issue, a cross-task feature calibration process is introduced in the second training stage, by which image embeddings from task-specific spaces are adaptively mapped to a unified task-shared feature space and the mapped image embeddings are expected to be better separated both within and across tasks in the task-shared space. This is achieved by the novel application of Mixture of Projectors (MoP). MoP can be considered as a modified Mixture of Experts (MoE)~\cite{MoE}, consisting of a set of independent and parallel projectors $\{P_m|=1,...,M\}$ and a gating network $G$. Each projector $P_m$ is a light-weight sub-network (\textit{e.g.}, a two-layer perceptron), and the gating network $G$ here is simply a linear layer followed by a softmax operator $\sigma(\cdot)$. Given any input image x and the adapted image encoder $f_t$ for the $t$-th task, let $\mathbf{W}_g \in \mathbb{R}^{M \times D}$ denote the linear layer of the gating network. Then, the output $\mathbf{g}_t \in \mathbb{R}^{M}$ of the gating network is
\begin{equation}
    \mathbf{g}_t=\sigma(\mathbf{W}_g f_t(\mathbf{x})) \,.
\end{equation}
With the learnable gating network and $M$ projectors, the original output from the adapted image encoder $f_t$ is adjusted adaptively as (\Cref{fig:method}, lower right)
\begin{equation}\label{eq:MoP}
    \mathbf{z}_t = f_t(\mathbf{x}) + \sum_{m=1}^M \mathbf{g}_{t,m}  P_m(f_t(\mathbf{x})) \,,
\end{equation}
where $\mathbf{z}_t$ is the calibrated feature output by MoP, and the output dimension (\textit{i.e.}, $D \times 1$) of each projector is the same as that of its input $f_t(\mathbf{x})$. From~\Cref{eq:MoP}, it is clear that, $\{\mathbf{g}_t\}$ are used to control the weights of mapping combinations. For different tasks, different gating scores distributions are learned. Through different mapping combinations, MoP can unify all task-specific spaces into a shared space. Ideally, such task-specific feature adjustment would lead to image embeddings in one space which are more comparable and separated among different classes and alleviate the image-text misalignment issue, if MoP is trained properly.  

In order to train the task-shared MoP, training data of certain form from all the learned tasks are necessary. However, in the exemplar-free CIL scenario, samples of old classes are not available, making it impossible to obtain the real image embeddings of old classes. Inspired by the relevant study~\cite{SLCA} which approximates the feature distribution of each class by a multivariate Gaussian distribution in the feature space, the feature distribution of the $c$-th class from the $t$-th task
is approximated by a Gaussian distribution $\mathcal{N}(\bm{\mu}_{t,c}, \mathbf{\Sigma}_{t,c})$, where $\bm{\mu}_{t,c}$ and $\mathbf{\Sigma}_{t,c}$  are, respectively, the mean and the covariance matrix of image embeddings of all original training images from the $c$-th class in the $t$-th task. $\bm{\mu}_{t,c}$ and $\mathbf{\Sigma}_{t,c}$ are obtained and saved after the model finishes Stage-I training on $t$-th task. Based on all distributions from task $1$ to task $t$, we sample $N_p$ pseudo image embeddings from each class's distribution to train the MoP with cross-entropy loss $\mathcal{L}_{ce}$.  


\subsection{Inference Strategy with Prediction Uncertainty}

Suppose the model has continually learned $T$ tasks, and the model is applied to predict the class of one test image whose class can be from any of the $T$ tasks. The test image is fed into $T$ adapted image encoders and then processed by the MoP module, and $T$ calibrated features $\{\mathbf{z}_t|t=1,...T\}$ are obtained. Now the key challenge is to select the most appropriate one, \textit{i.e.}, the one that is from the encoder associated with the task of the image, from the $T$ calibrated features for prediction. Here a novel feature selection strategy is proposed based on the MoP's prediction uncertainty.

A model often shows uncertain predictions for those out-of-distribution images during model inference~\cite{malinin2018predictive}. When visual embeddings deviated from the estimated Gaussian distributions of all learned classes that are used to train the MoP, the corresponding MoP outputs more likely lead to higher model prediction uncertainties compared to those visual embeddings from high-density region of these class Gaussian distributions. We expect that for any test image from the learned $t$-th task, the model prediction uncertainty is more likely lower when using the adapted image encoder of task $t$ to extract image embedding than using that of any other task, because this image embedding more likely falls into the high-density region of the Gaussian distribution associated with the class of the input image, and meanwhile deviates from Gaussian distributions of other classes. Here the prediction uncertainty for any input image is measured by the entropy. For each $\mathbf{z}_t$, we calculate its similarities with all $C$ class text embeddings and obtain the output probability distribution $\mathbf{p} = [p_1, p_2, \ldots, p_{C}]^{\top}$. Then the entropy is computed as 
\begin{equation}\label{eq:entropy}
    \mathcal{E}_t(\mathbf{p}) = - \sum_{c=1}^{C} p_c \log p_c \,, 
\end{equation}

Finally, we select the calibrated feature with the lowest entropy for classification by computing the similarities with text embeddings of all $C$ classes, and the class with the highest similarity is adopted as the final prediction.

\begin{table*}[t]
  \centering
  \footnotesize
  \renewcommand{\arraystretch}{0.8}
  \caption{Performance of different methods on CIFAR100 and ImageNet-R under the 5-, 10-, 20-task settings. ``Memory'' means whether the method requires old class samples. The subscript number represents the standard deviation.}
  \label{tab:cifar-ir}
  \setlength{\tabcolsep}{3pt}
  \begin{tabular}{cccccccccccccc}
    \toprule
    \multirow{5}{*}{Method} & \multirow{5}{*}{Memory} & \multicolumn{6}{c}{CIFAR100} & \multicolumn{6}{c}{ImageNet-R}  \\
    
    \cmidrule(lr){3-14}
    
     & & \multicolumn{2}{c}{5-task} & \multicolumn{2}{c}{10-task} & \multicolumn{2}{c}{20-task} &  \multicolumn{2}{c}{5-task} & \multicolumn{2}{c}{10-task} & \multicolumn{2}{c}{20-task} \\

    \cmidrule(lr){3-4}\cmidrule(lr){5-6}\cmidrule(lr){7-8}\cmidrule(lr){9-10}\cmidrule(lr){11-12}\cmidrule(lr){13-14}
    
     & & \textit{Last-A} & \textit{Avg-A} & \textit{Last-A} & \textit{Avg-A} & \textit{Last-A} & \textit{Avg-A} & \textit{Last-M} & \textit{Avg-M} & \textit{Last-M} & \textit{Avg-M} & \textit{Last-M} & \textit{Avg-M} \\
     
    \midrule 
    
    PROOF \cite{PROOF} & \Checkmark & 76.59\std{0.28} & 84.21\std{0.08} & 75.90\std{0.20} & 84.80\std{0.06} & 75.31\std{0.26} & 85.06\std{0.14} & 78.19\std{0.16} & 83.73\std{0.19} & 77.39\std{0.09} & 84.26\std{0.68} & 76.27\std{0.36} & 84.19\std{0.09} \\
    
    PR-CLIP \cite{PR-CLIP} & \Checkmark & 76.13\std{0.35} & 83.46\std{0.24} & 75.43\std{0.89} & 84.32\std{0.56} & 75.16\std{0.81} & 84.52\std{0.35} & 76.80\std{0.08} & 81.52\std{0.86} & 76.40\std{0.08} & 82.35\std{0.64} & 76.05\std{0.12} & 82.81\std{0.61} \\
    
    \cmidrule(lr){1-14}

    L2P \cite{L2P} & \XSolidBrush & 76.13\std{0.03} & 84.16\std{0.17} & 73.01\std{0.12} & 82.17\std{0.11} & 68.03\std{0.30} & 79.59\std{0.17} & 75.55\std{0.11} & 82.32\std{0.09} & 72.86\std{0.28} & 81.10\std{0.25} & 67.94\std{0.19} & 78.28\std{0.21} \\
    
    DualPrompt \cite{DualPrompt} & \XSolidBrush & 77.07\std{0.44} & 85.31\std{0.17} & 74.12\std{0.15} & 82.98\std{0.08} & 69.66\std{0.73} & 80.28\std{0.32} & 76.94\std{0.31} & 83.08\std{0.02} & 73.76\std{0.14} & 82.08\std{0.03} & 69.30\std{0.30} & 78.85\std{0.49} \\
    
    CODA-Prompt \cite{CODA-Prompt} & \XSolidBrush & 76.37\std{0.59} & 84.65\std{0.06} & 68.76\std{0.21} & 81.18\std{0.36} & 56.21\std{2.41} & 74.65\std{0.55} & 72.47\std{0.33} & 80.79\std{0.09} & 65.54\std{0.66} & 77.67\std{0.27} & 62.62\std{0.83} & 76.43\std{0.25} \\

    EASE \cite{EASE} & \XSolidBrush & 80.70\std{0.18} & 85.27\std{0.09} & 77.58\std{0.02} & 85.68\std{0.15} & 73.95\std{0.12} & 84.09\std{0.06} & 76.72\std{0.12} & 83.77\std{0.06} & 75.38\std{0.17} & 83.71\std{0.13} & 73.68\std{0.18} & 82.15\std{0.04} \\

    SLCA \cite{SLCA} & \XSolidBrush & 80.26\std{0.14} & 87.35\std{0.07} & 73.39\std{0.38} & 84.46\std{0.21} & 69.00\std{0.49} & 82.02\std{0.34} & 79.54\std{0.18} & 85.29\std{0.04} & 76.21\std{0.51} & 84.79\std{0.04} & 73.54\std{0.33} & 83.80\std{0.11} \\
    
    
    AttriCLIP \cite{AttriCLIP} & \XSolidBrush & 68.35\std{0.28} & 79.39\std{0.06} &  66.0\std{1.41} & 78.84\std{0.09} & 64.70\std{0.89} & 78.19\std{0.26} & 74.97\std{0.11} & 81.44\std{0.14} & 72.71\std{0.49} & 81.36\std{0.16} & 72.67\std{1.17} & 81.28\std{0.25} \\
    
    MoE-Adapters \cite{MoE-adapters} & \XSolidBrush & 78.13\std{0.70} & 85.27\std{0.16} & 76.33\std{0.37} & 84.75\std{0.28} & 75.27\std{0.52} & 84.34\std{0.30} & 78.68\std{0.38} & 83.61\std{0.21} & 78.54\std{0.35} & 84.15\std{0.15} & 77.95\std{0.31} & 83.78\std{0.04} \\

    Lang-CL \cite{Lang-CL} & \XSolidBrush & 83.84\std{0.17} & 89.44\std{0.11} & 81.49\std{0.15} & 88.44\std{0.20} & 77.26\std{0.59} & 86.40\std{0.07} & 82.46\std{0.31} & 87.14\std{0.42} & 81.77\std{0.12} & 87.65\std{0.29} & 79.32\std{0.23} & 86.77\std{0.23} \\

    \cmidrule(lr){1-14}
    
    Ours & \XSolidBrush & \textbf{84.43\std{0.21}} & \textbf{90.13\std{0.13}} & \textbf{82.52\std{0.36}} & \textbf{89.60\std{0.04}} & \textbf{77.68\std{0.52}} & \textbf{87.11\std{0.37}} & \textbf{84.02\std{0.28}} & \textbf{88.42\std{0.34}} & \textbf{83.58\std{0.43}} & \textbf{88.60\std{0.26}} & \textbf{82.24\std{0.39}} & \textbf{88.21\std{0.16}}  \\
    
    \bottomrule
  \end{tabular}

\end{table*}

\section{Experiments}

\subsection{Experimental Setup}

\begin{table*}[t]
    \footnotesize
    \centering
    \renewcommand{\arraystretch}{0.8}
    \caption{Performance of different methods on Cars196 and Skin40. “Memory” means whether the method requires old class samples. The subscript number represents the standard deviation.}
    \label{tab:car_skin}
    \setlength{\tabcolsep}{5.5pt}
    \begin{tabular}{cccccccccccc}
        \toprule
        \multirow{4}{*}{Method} & \multirow{4}{*}{Memory}  & \multicolumn{6}{c}{Cars196} & \multicolumn{4}{c}{Skin40}  \\ 
        
        \cmidrule(l){3-12} 
        &   & \multicolumn{2}{c}{5-task} & \multicolumn{2}{c}{10-task} & \multicolumn{2}{c}{20-task} & \multicolumn{2}{c}{5-task} & \multicolumn{2}{c}{10-task} \\ 
        
        \cmidrule(l){3-4} \cmidrule(l){5-6} \cmidrule(l){7-8} \cmidrule(l){9-10} \cmidrule(l){11-12}
        
         &  & \textit{Last-M}  & \textit{Avg-M} & \textit{Last-M} & \textit{Avg-M} & \textit{Last-M} & \textit{Avg-M} & \textit{Last-A} & \textit{Avg-A} & \textit{Last-A} & \textit{Avg-A} \\ 
         
        \midrule

        PROOF \cite{PROOF} & \Checkmark & 78.33\std{0.14} & 87.03\std{0.28} & 79.85\std{0.67} & 87.8\std{0.53} & 79.09\std{0.32} & 88.42\std{0.13} & 53.00\std{0.35} & 73.00\std{0.62} & 50.08\std{0.13} & 71.07\std{0.18} \\

        PR-CLIP \cite{PR-CLIP} & \Checkmark & 69.96\std{0.21} & 82.93\std{0.35} & 71.17\std{1.18} & 83.55\std{0.17} & 70.62\std{0.41} & 84.10\std{0.60} & 51.83\std{2.10} & 67.09\std{1.88} & 50.08\std{1.91} & 65.25\std{2.59} \\

        \cmidrule(lr){1-12}

        L2P \cite{L2P}  & \XSolidBrush  & 71.29\std{0.87}  & 81.68\std{0.52}  & 66.40\std{0.90} & 78.71\std{0.98} & 56.12\std{1.13} & 72.50\std{0.10} & 43.92\std{0.63} & 66.57\std{0.97} & 37.25\std{1.25} & 58.69\std{0.78}  \\
        
        DualPrompt \cite{DualPrompt} & \XSolidBrush  & 71.15\std{0.74} & 81.82\std{0.55} & 66.35\std{0.28} & 79.07\std{0.77} & 56.75\std{0.50} & 73.35\std{1.39} & 42.17\std{1.28} & 66.97\std{0.75} & 37.50\std{1.32} & 61.26\std{0.88}  \\
        
        CODA-Prompt \cite{CODA-Prompt} & \XSolidBrush  & 71.47\std{1.07} & 81.69\std{0.28}  & 67.75\std{3.01} & 80.52\std{1.49} & 55.36\std{0.68} & 72.07\std{1.51} & 40.17\std{0.88} & 62.37\std{0.34} & 30.25\std{2.38} & 54.84\std{1.06}  \\\
        
        EASE \cite{EASE} & \XSolidBrush  & 66.41\std{1.25}  & 76.88\std{0.53} & 65.88\std{1.32} & 77.32\std{1.80} & 60.74\std{3.26} & 74.90\std{0.46} & 53.50\std{0.87} & 69.85\std{0.53} & 50.75\std{0.66} & 70.62\std{0.31} \\
        
        SLCA \cite{SLCA} & \XSolidBrush  & 76.44\std{0.85}  & 86.01\std{0.84} & 66.82\std{2.23}  & 81.05\std{0.92} & 52.82\std{1.78} & 74.31\std{0.98} & 50.00\std{1.15} & 71.56\std{0.78} & 46.75\std{0.87} & 66.51\std{1.24} \\
        
         
        
        MoE-Adapters \cite{MoE-adapters}  & \XSolidBrush & 71.67\std{1.21}  & 83.58\std{1.01}  & 67.30\std{2.47} & 81.34\std{1.32} & 61.72\std{3.01} & 78.78\std{0.34} & 41.33\std{1.38} & 66.09\std{1.55} & 23.75\std{1.52} & 52.44\std{0.78} \\
        
        Lang-CL \cite{Lang-CL}  & \XSolidBrush   & 79.00\std{0.39}  & 87.72\std{0.19}  & 78.07\std{0.22} & 87.79\std{0.11} & 77.34\std{0.39} & 86.74\std{0.16} & 54.67\std{0.29} & 72.89\std{0.15} & 51.42\std{1.23} & 69.12\std{0.89} \\

        \cmidrule(lr){1-12}
        
        Ours  & \XSolidBrush  & \textbf{83.10\std{1.15}}  & \textbf{90.08\std{0.83}}  & \textbf{81.52\std{0.63}}  & \textbf{89.13\std{0.72}}  & \textbf{79.96\std{0.81}}  & \textbf{89.15\std{0.72}} & \textbf{55.83\std{0.76}} & \textbf{73.07\std{0.33}} & \textbf{54.50\std{1.32}} & \textbf{71.33\std{0.48}} \\ 
        
        \bottomrule
    \end{tabular}

\end{table*}

\noindent \textbf{Datasets:} We evaluated our method on five datasets: CIFAR100 \cite{cifar100}, ImageNet-R \cite{imagenet_r}, Cars196 \cite{cars196} for natural scenarios, and Skin40 \cite{Skin40} for medical applications. Each dataset includes a training and a test set, encompassing 100 classes for CIFAR100, 200 classes for ImageNet-R, and 196 and 40 classes for Cars196 and Skin40, respectively. For CIFAR100, ImageNet-R, and Cars196, we performed experiments in 5-task, 10-task, and 20-task settings, while Skin40 was evaluated in 5-task and 10-task settings only.


\noindent \textbf{Implementation details:} We utilize the OpenAI CLIP ViT-B/16 model \cite{ViT} as the backbone in our experiments. The AdamW optimizer with a weight decay of 0.0001 and cosine learning rate decay is employed. The model is trained for 30 epochs per task with a learning rate of 0.001. The hidden dimension of the task-specific adapter $r$ is set to 64, and $M$ is set to 3. During Stage-II, we sample $N_p = 256$ pseudo features per class distribution, with MoP trained for 5 epochs. Each experiment is conducted three times using different random seeds to ensure robustness of the results.


\noindent \textbf{Evaluation metrics:} For class-balanced datasets (CIFAR100, Mini-ImageNet100, Skin40), we use last accuracy (\textit{Last-A}) and average accuracy (\textit{Avg-A}). \textit{Last-A} measures performance on all classes after the final task; \textit{Avg-A} is the mean accuracy across all tasks. For class-imbalanced datasets (ImageNet-R, Cars196), we report last mean class recall (\textit{Last-M}) and average mean class recall (\textit{Avg-M}), where MCR is the average recall across all classes.


\subsection{Effectiveness Evaluation}

To evaluate the effectiveness of our method and demonstrate its superiority, we compare our method with several CIL baselines. These baselines includes L2P~\cite{L2P}, DualPrompt~\cite{DualPrompt}, CODA-Prompt~\cite{CODA-Prompt}, EASE~\cite{EASE}, SLCA~\cite{SLCA}, MoE-Adapters~\cite{MoE-adapters}, AttriCLIP~\cite{AttriCLIP}, Lang-CL \cite{Lang-CL}, PROOF \cite{PROOF}, PR-CLIP \cite{PR-CLIP}. It is important to note that both PROOF and PR-CLIP requires the use of old class samples for replay. we allocate a fixed memory buffer with sizes are set as follows: 2000 samples for CIFAR100, 4000 for ImageNet-R, 400 for Cars196, and 80 for Skin40. 

\Cref{tab:cifar-ir} demonstrates a comparison of our method’s performance against strong baselines on CIFAR100 and ImageNet-R, where we achieve state-of-the-art results. Specifically, on CIFAR100, our approach improves the \textit{Last-A} score by 6.62\% and 7.09\% over memory-based methods like PROOF and PR-CLIP in the 10-task setting. For ImageNet-R, our method consistently outperforms typical prompt-based methods, such as L2P, DualPrompt, and CODA-Prompt, by at least 10\% in \textit{Last-M} for the 20-task setting.

To further evaluate performance on fine-grained datasets with confusable classes, we experimented with Cars196 and Skin40. As shown in \Cref{tab:car_skin}, our method excels on both datasets, particularly on Cars196, where we outperform task-specific competitors like EASE and Lang-CL by 15.64\% and 3.45\%, respectively, in the 10-task setting. Besides, the outstanding performance of our method on Skin40, a medical dataset, demonstrates its superior generalization ability.

Overall, the results across these datasets validate the effectiveness of our method, particularly in handling confusable classes and cross-domain data. 

\begin{table}[t]
    \footnotesize
    \centering
    \renewcommand{\arraystretch}{0.8}
    \caption{Ablation study of our methods on CIFAR100 and ImageNet-R under the 10-task setting. ``TSA'': task-specific adapters; The subscript number represents the standard deviation.}
    \label{tab:ablation}
    \begin{tabular}{cccccc}
        \toprule
        \multirow{2}{*}{TSA} & \multirow{2}{*}{MoP}  & \multicolumn{2}{c}{CIFAR100} & \multicolumn{2}{c}{ImageNet-R} \\ 
        
        \cmidrule(l){3-6} 
        & & \textit{Last-A}  & \textit{Avg-A} & \textit{Last-M} & \textit{Avg-M} \\ 
         
        \midrule

        & & 63.77 & 76.42  & 72.80 & 79.34 \\
        \Checkmark & & 70.53\std{0.21} & 81.70\std{0.09} & 78.06\std{0.37} & 85.64\std{0.27} \\
        & \Checkmark & 76.18\std{0.19} & 84.58\std{0.31} & 79.68\std{0.44} & 85.08\std{0.26} \\
        \Checkmark & \Checkmark & 82.52\std{0.36} & 89.60\std{0.04} & 83.58\std{0.43} & 88.60\std{0.26} \\
        \bottomrule
    \end{tabular}
\end{table}

\subsection{Ablation Study}
In this section, we conduct ablation analysis to investigate the effectiveness of key components of our proposed method for CIL. As shown in \Cref{tab:ablation}, Row 1 stands for baseline which is CLIP zero-shot. The addition of TSA (row 2) improves the performance by more than 5\% in \textit{Last-A/M} on both datasets compared to the baseline, indicating that using task-specific adapters for task adaptation is effective. The participation of MoP only (row 3) which is trained using pseudo-features derived from class Gaussian distributions that are constructed using pre-trained image embeddings performs better than baseline, demonstrating the effectiveness of our proposed cross-task representation calibration even with pre-trained image features. When MoP is combined with TSA and the inference is performed based on entropy (row 4), the model outperforms that of either component when used individually (row 2 or row 3), highlighting that our proposed two-stage training can better help cross-task class separation. In summary, the ablation results verify that each component in our method boosts the performance.

\begin{figure}[t]
  \centering
  \includegraphics[width=0.49\columnwidth]{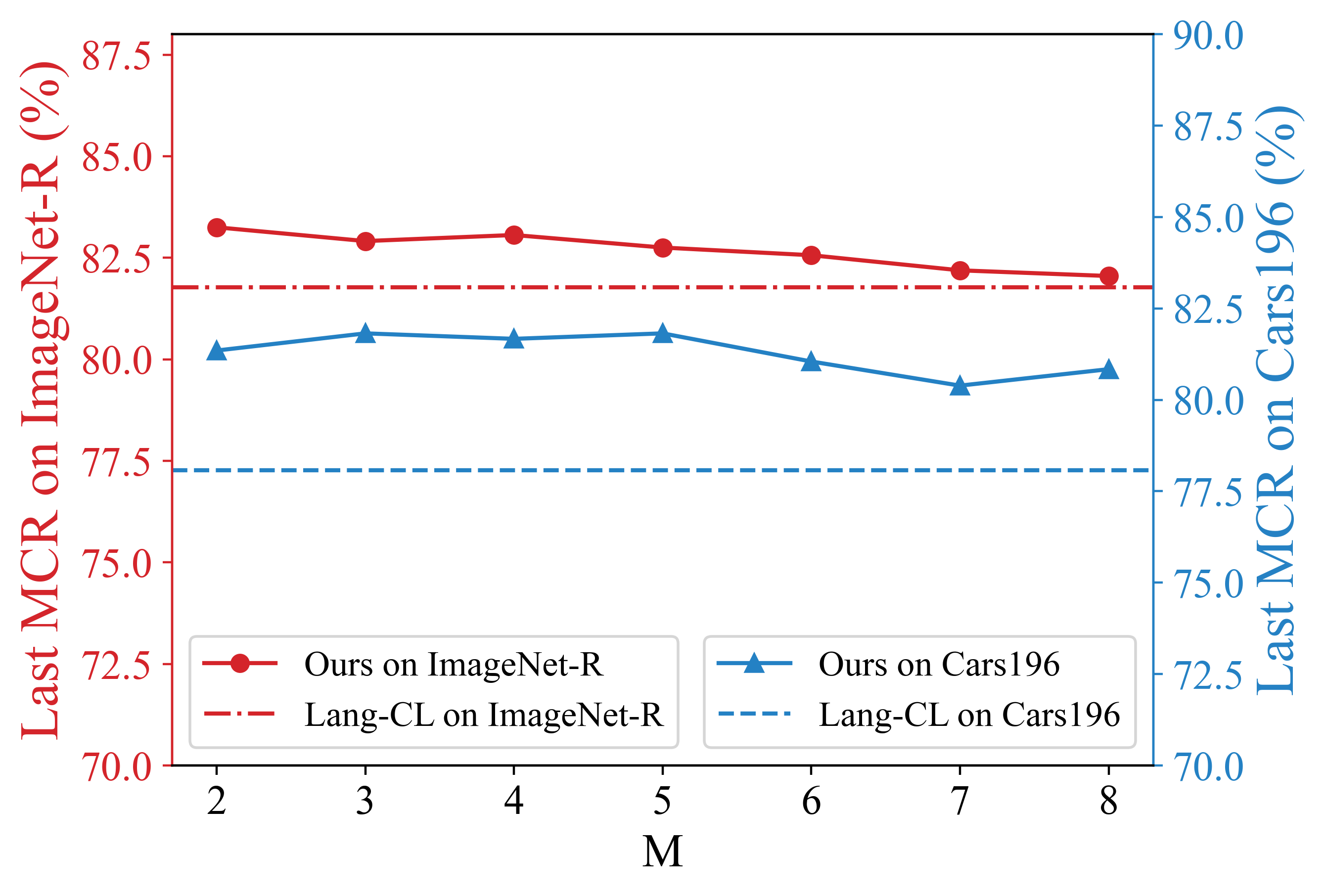}
  \includegraphics[width=0.49\columnwidth]{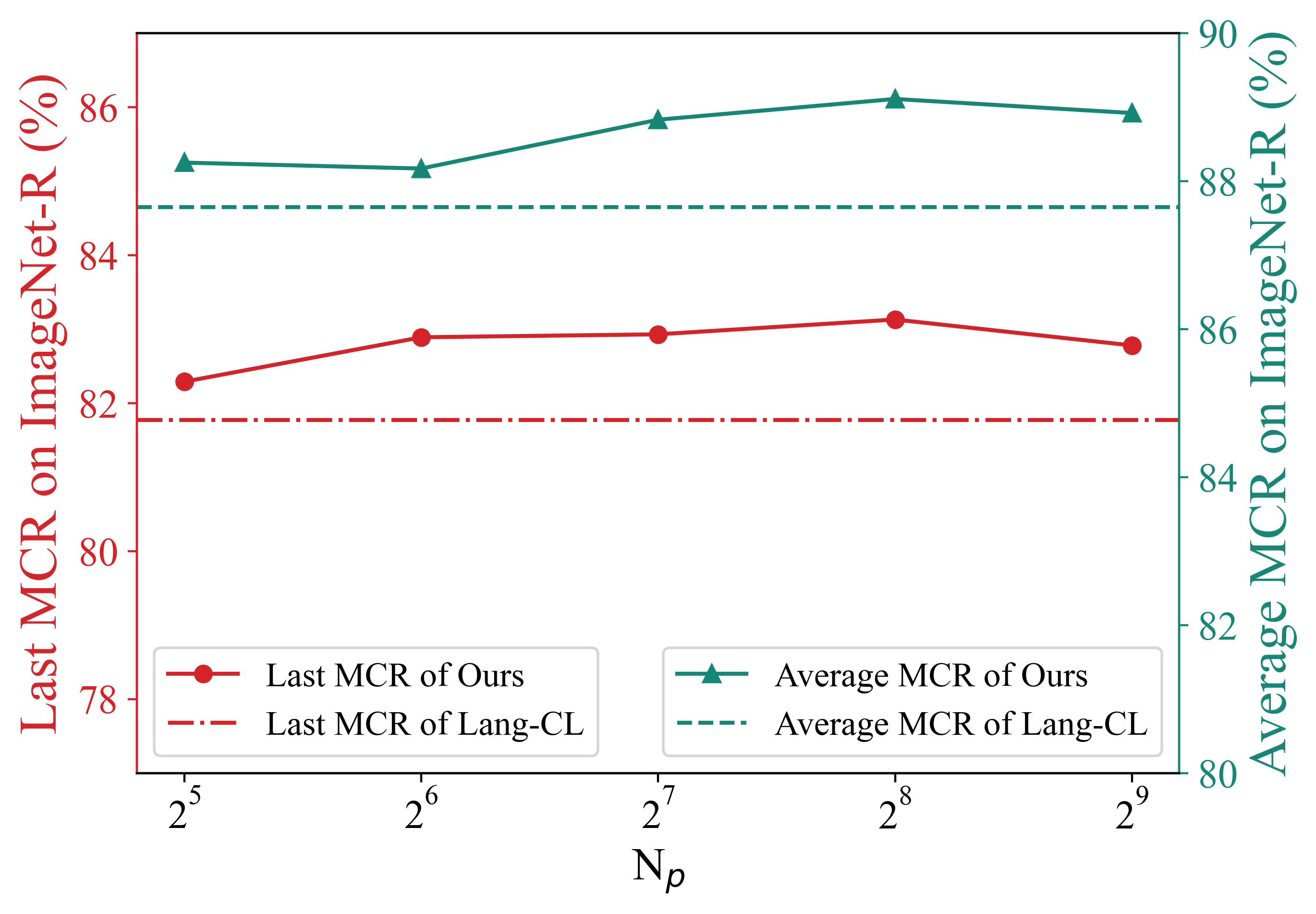}
  \caption{Sensitivity analysis of $M$ (Left) and $\boldsymbol{N_p}$ (Right) on ImageNet-R (red) and Cars196 (blue) under the 10-task setting.}
  \label{fig:M&N}
\end{figure}




\begin{figure}[t]
  \centering
  \includegraphics[width=0.47\columnwidth]{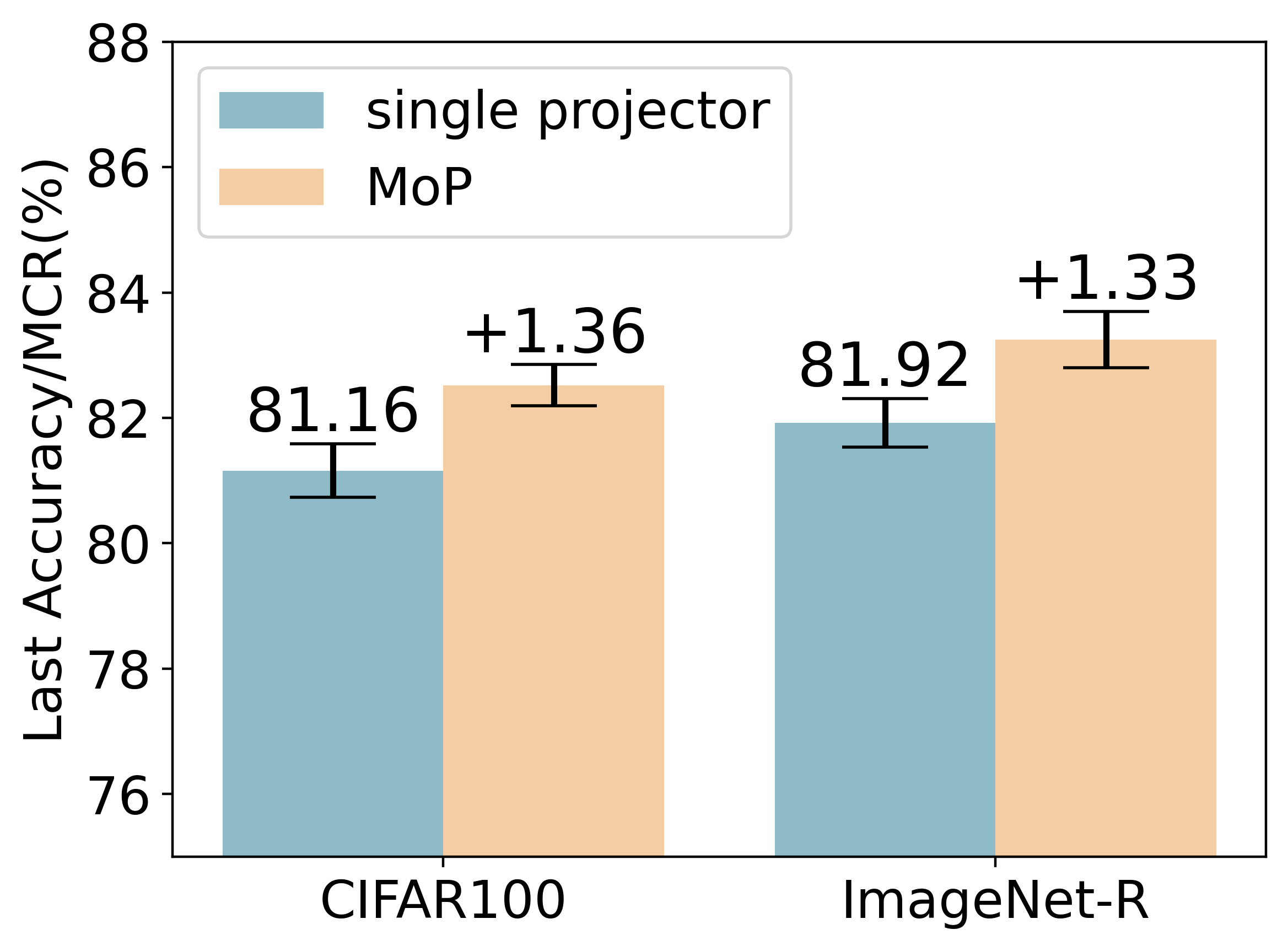}
  \includegraphics[width=0.47\columnwidth]{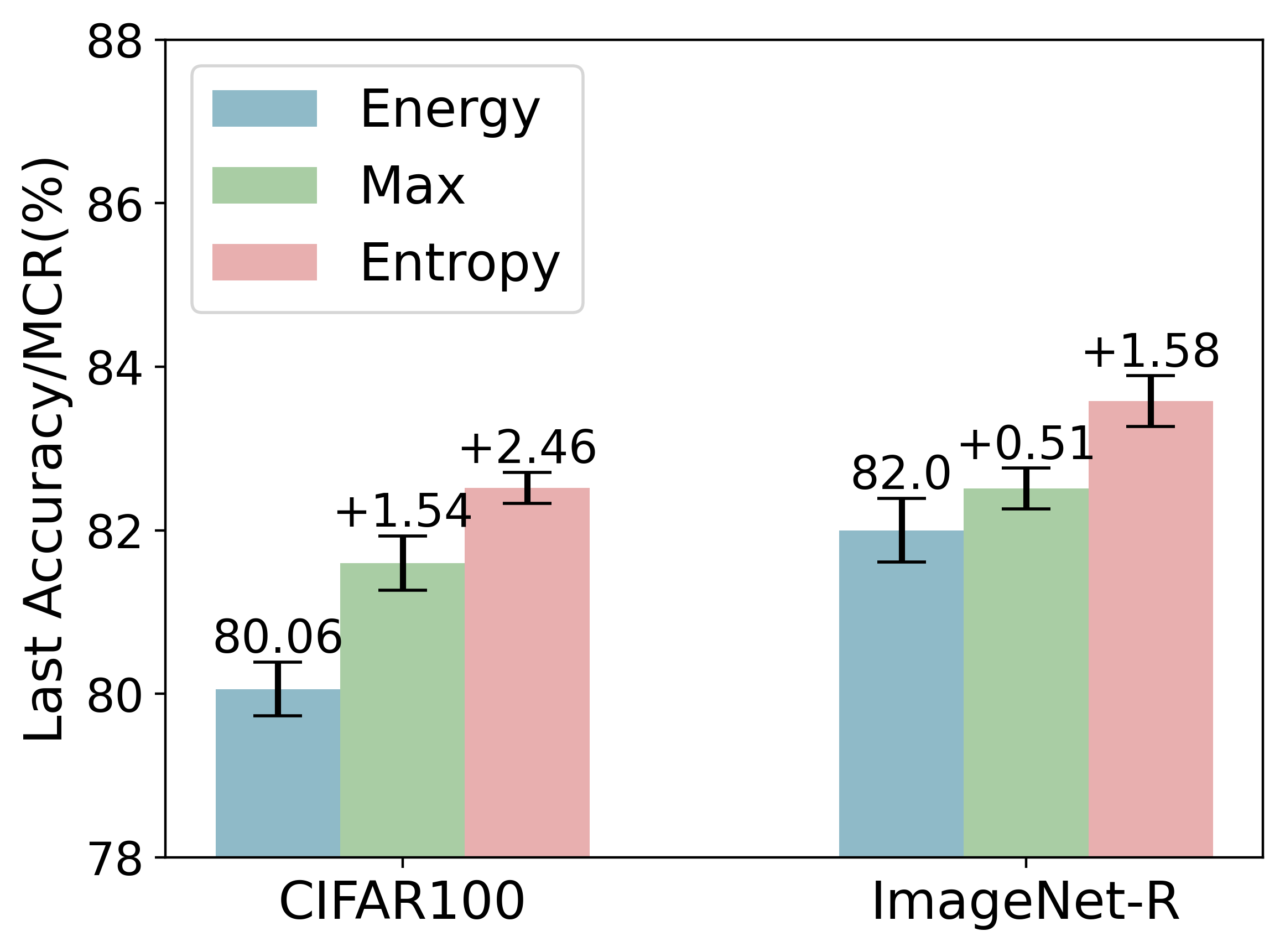}
  \caption{Further analysis of MoP (Left) and the inference strategy (Right) on CIFAR100 and ImageNet-R. The leftest bar is set as baseline in each group and the numbers on the other bars represent the relative changes compared to the baseline. The black error bar represents standard deviation.
    }
  \label{fig:proj&inference-analysis}
\end{figure}
\subsection{Sensitivity Analysis}

We analyze the sensitivity of our method to the number of projectors $M$ in the MoP and the number of pseudo-features sampled per class $N_p$. As shown in \Cref{fig:M&N} (left), our method remains stable and outperforms Lang-CL on ImageNet-R and Cars196 when $M$ is set between 2 and 8. Only when $M > 5$ does performance slightly decline, likely due to overfitting from the increased number of parameters. Similarly, as shown in \Cref{fig:M&N} (right), our method is also robust to different values of $N_p$. Performance only drops when $N_p$ is too small (insufficient diversity) or too large (possible overfitting). Overall, our method is not sensitive to moderate changes in $M$ or $N_p$.

\subsection{Further Analysis}

\noindent \textbf{MoP or single projector:}
We compared the performance of MoP with that of a single projector using in our method to evaluate the effectiveness of MoP. As illustrated on the left of \Cref{fig:proj&inference-analysis}, employing a gating network to integrate multiple projectors results in a performance improvement of over 1\% compared to a single projector across both datasets. These results demonstrate that a mapping module with greater capacity, facilitated by a gating network that dynamically provides mapping combinations based on the input, can more effectively achieve cross-task feature calibration.

\noindent \textbf{Different inference strategies:} We compared our inference strategy with other approaches at \Cref{fig:proj&inference-analysis}. Our inference strategy (\textit{Entropy}) outperforms the other approaches by a significant margin on both datasets. \textit{Max} means that selecting the highest probability from the probability distributions of the $T$ tasks (\textit{i.e.}, $T$ probability vectors of dimension $C$) as the predicted class. \textit{Energy} approach leverages the energy function commonly used in out-of-distribution (OOD) detection~\cite{energy}. Energy function is also a measurement of model's uncertainty. It computes energy from the logits before softmax normalization. However, in CLIP, since the cosine similarity scores range between -1 and 1, the energy differences between OOD and ID samples become minimal, leading to increased difficulty in distinguishing between them. So the \textit{energy} approach exhibits the poorest performance.

\section{Conclusion}

In this paper, we propose a novel VLM-based class-incremental learning framework utilizing a two-stage training approach. The first stage employs task-specific adapters to capture representations, while the second stage maps these features into a unified cross-task space using MoP module, reducing class confusion. Additionally, we introduce an inference strategy based on prediction uncertainty, selecting the calibrated feature with the lowest uncertainty to enhance accuracy. Our evaluations demonstrate that this two-stage strategy significantly improves continual learning performance, achieving state-of-the-art results across multiple benchmarks.




\bibliographystyle{IEEEbib}
\bibliography{icme2026references}

\clearpage
\appendix

\section*{Additional Experiments and Analysis}



\subsection{Additional Evaluation}

\begin{figure*}[t]
  \centering
  \includegraphics[width=0.6\columnwidth]{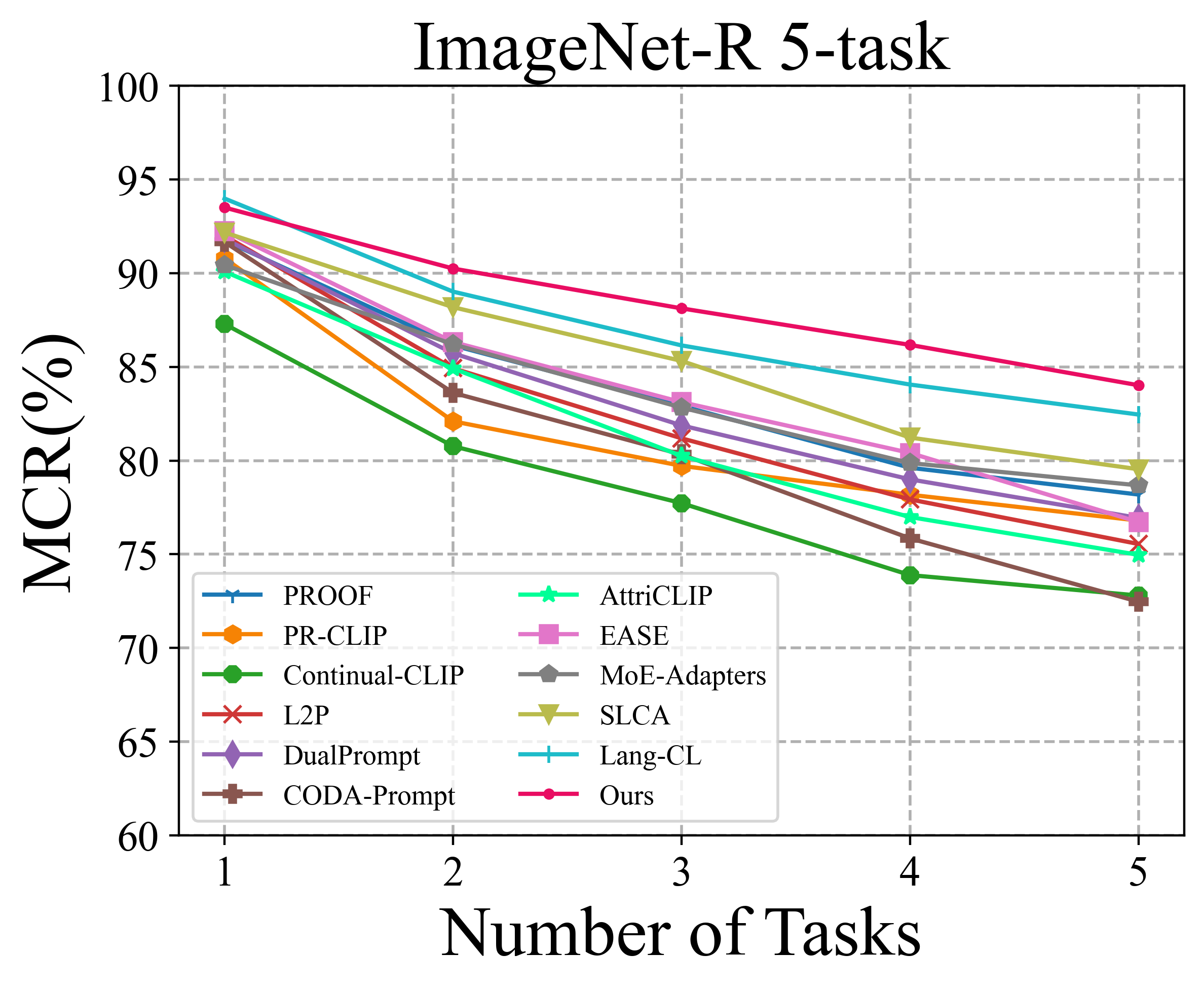}
  \includegraphics[width=0.6\columnwidth]{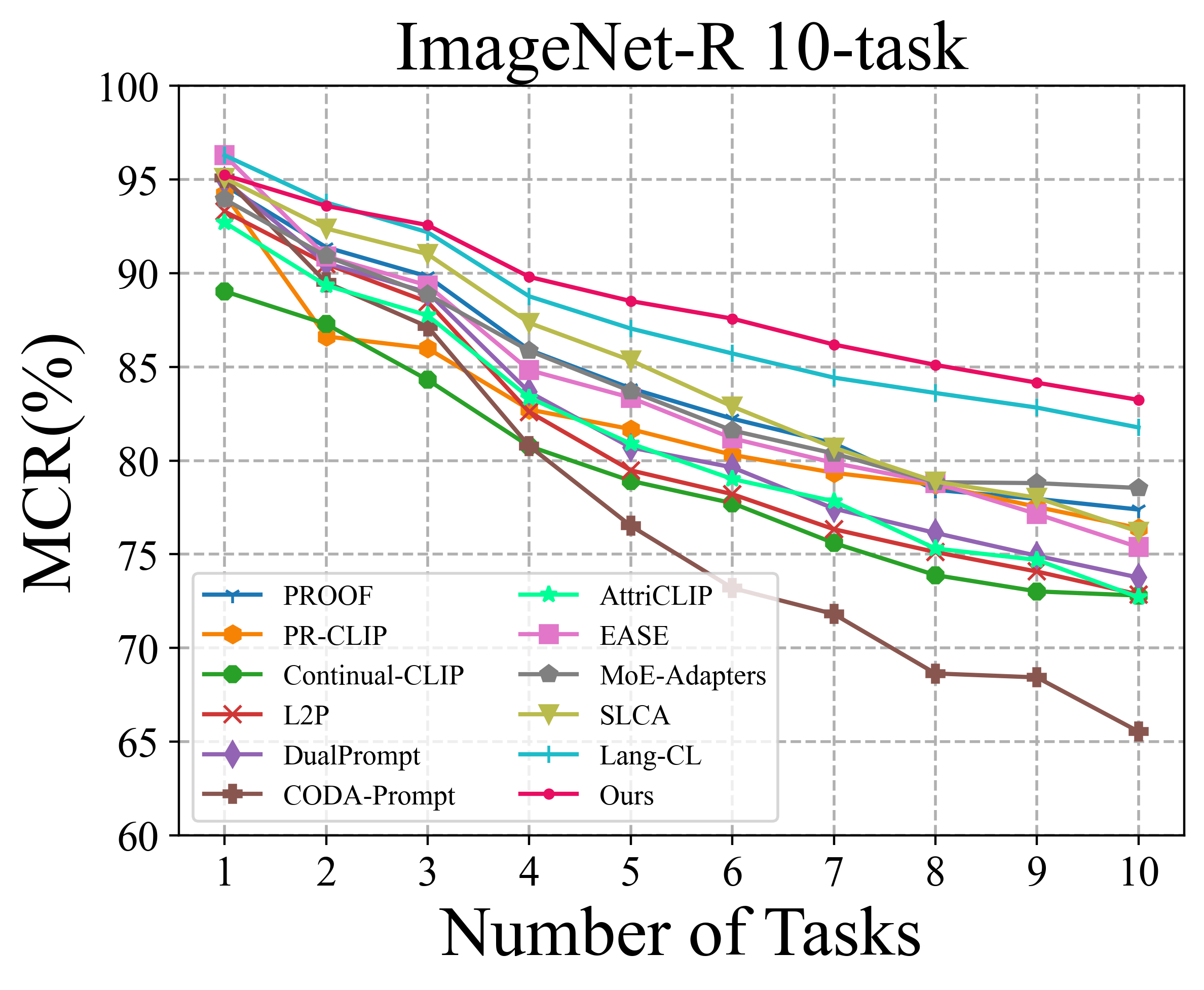}
  \includegraphics[width=0.6\columnwidth]{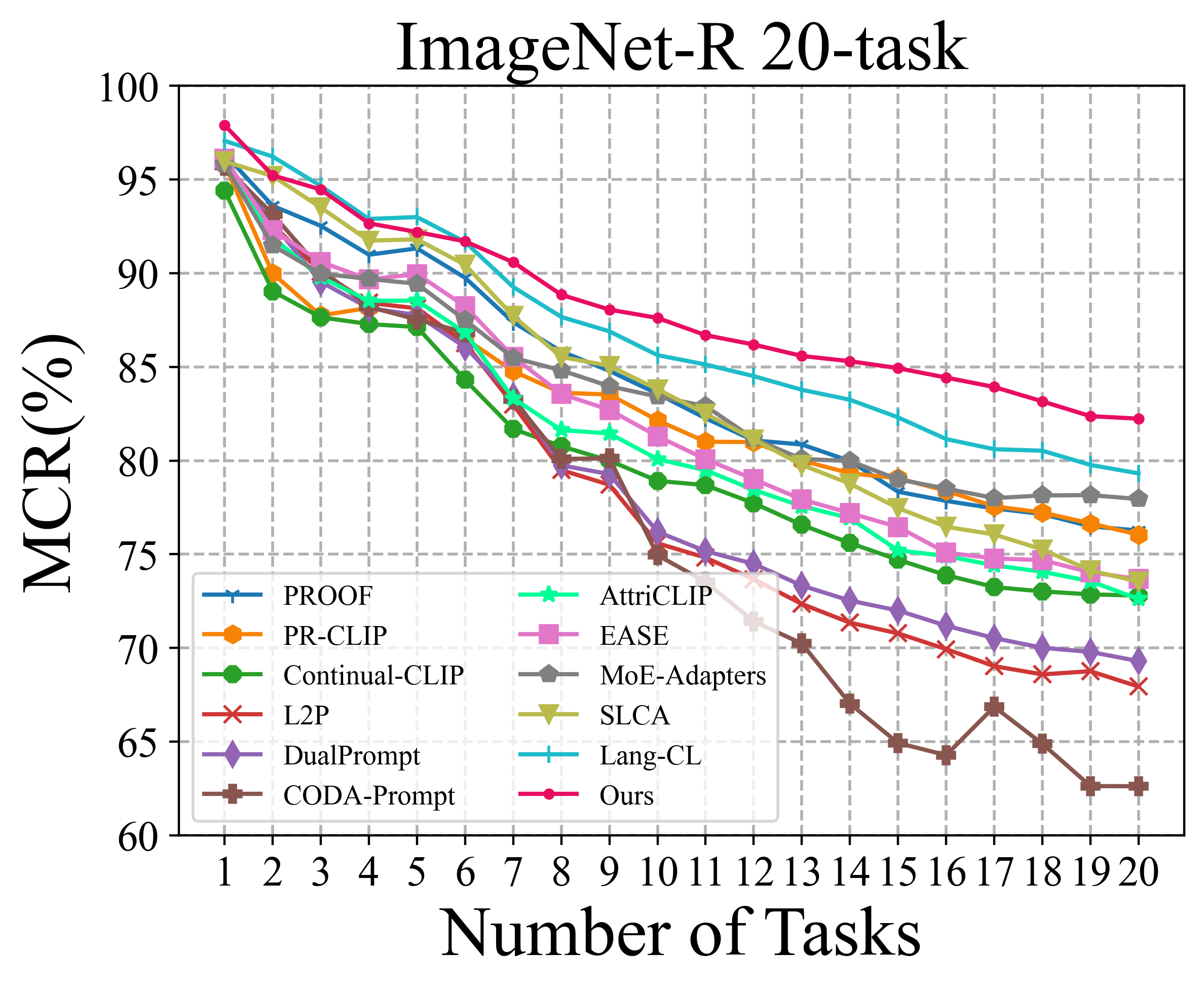}
  \\
  \includegraphics[width=0.6\columnwidth]{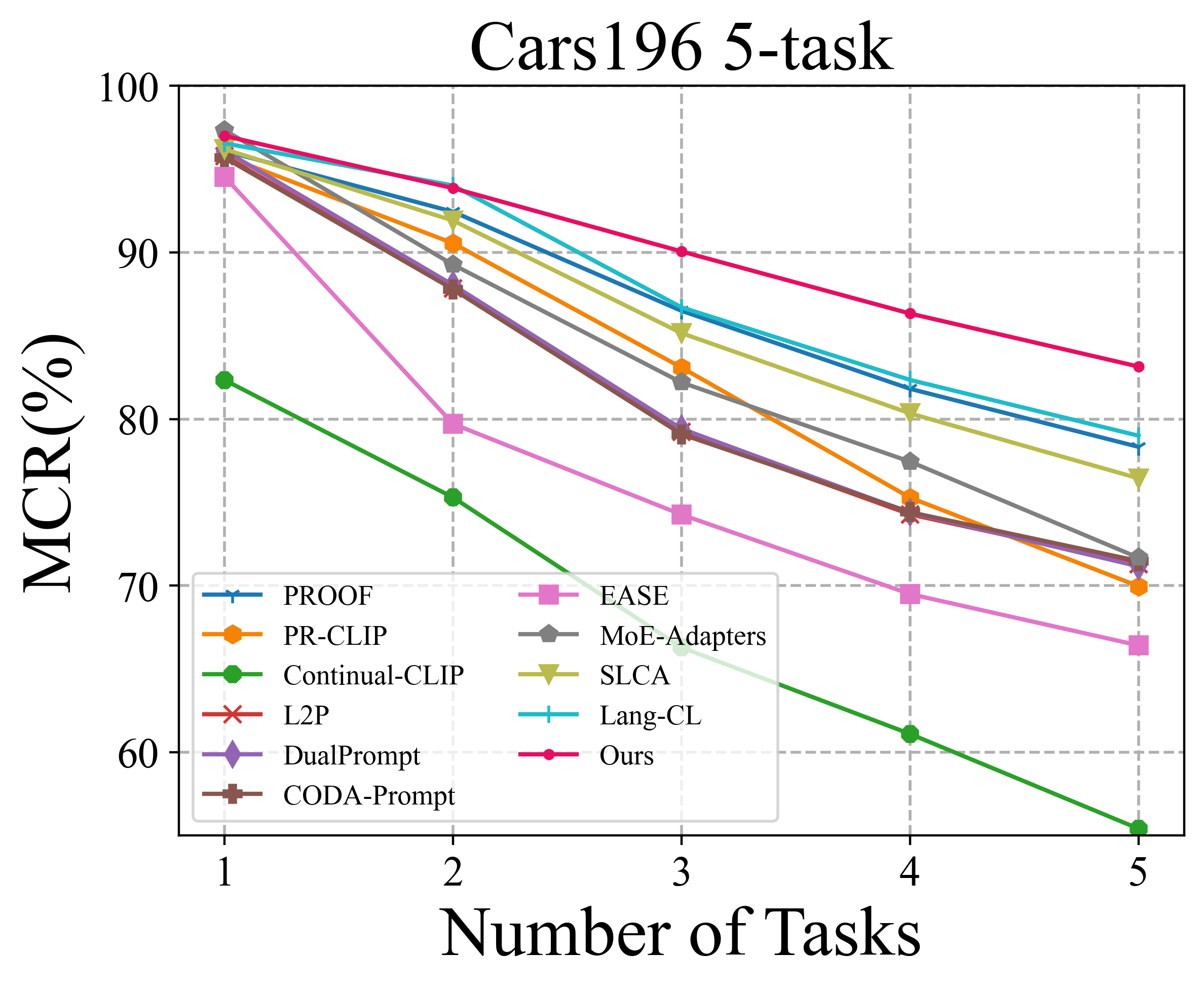}
  \includegraphics[width=0.6\columnwidth]{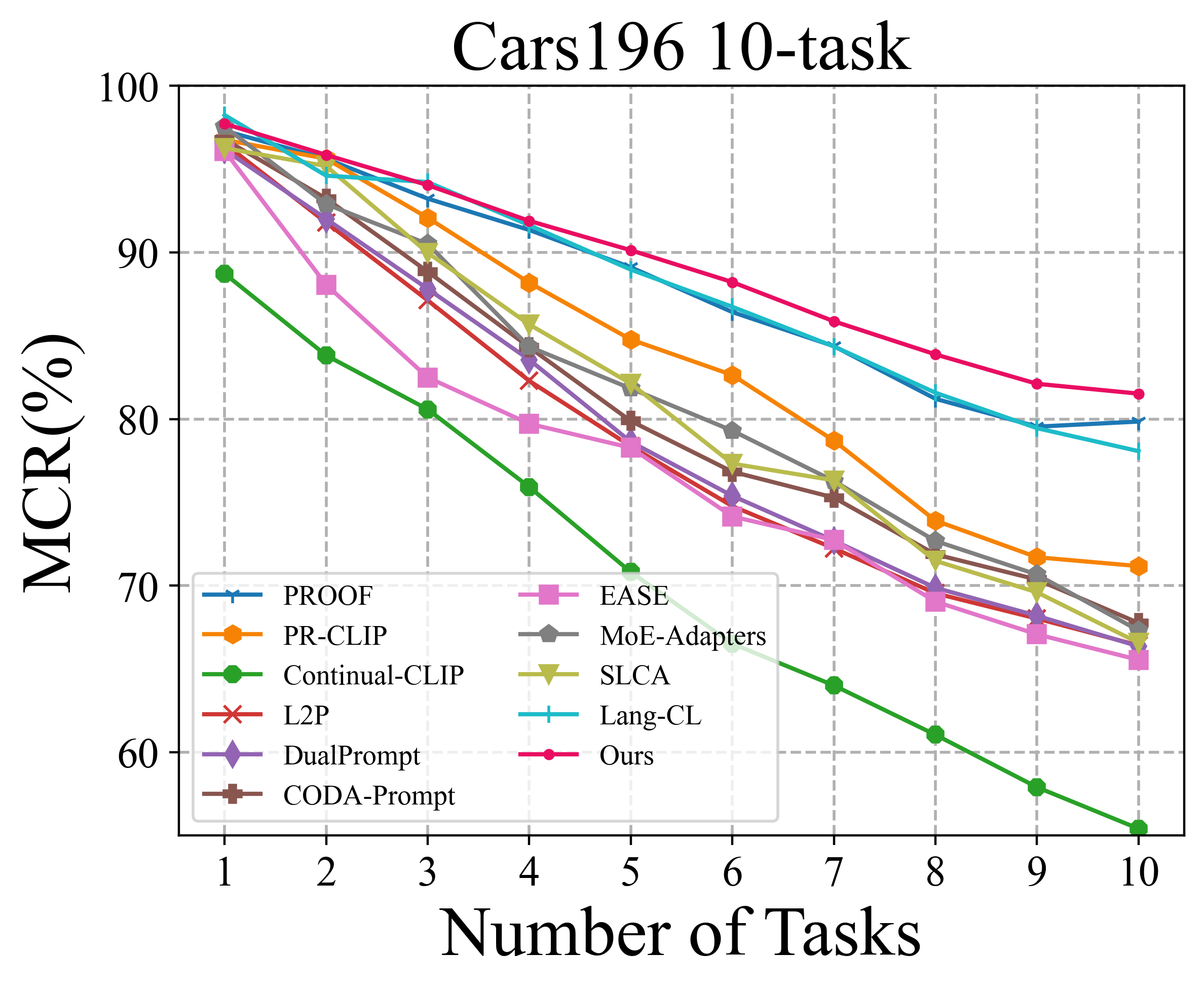}
  \includegraphics[width=0.6\columnwidth]{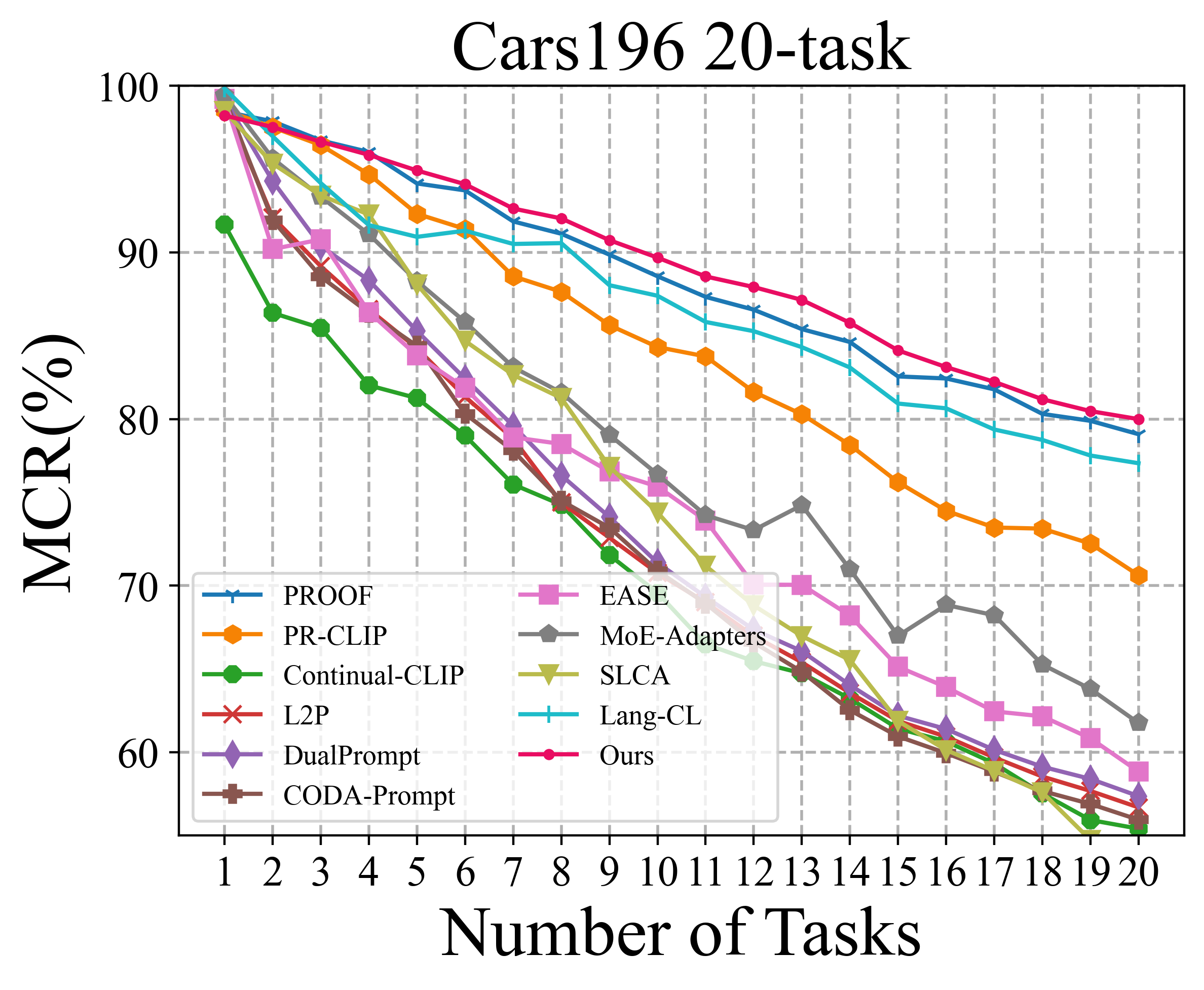}
  \caption{Performance of different methods over the whole continual learning process on ImageNet-R and Cars196 under the 5-, 10- and 20-task settings.}
  \label{fig:curve}
\end{figure*}

To further assess our method’s performance on datasets with confusable classes, we conducted experiments on Mini-ImageNet100 datasets. As shown in \Cref{tab:mini-IN}, due to the overlap between Mini-ImageNet100 and the CLIP pre-training data, CLIP achieves an impressive zero-shot performance of 89.99\%. As a result, the performance gaps among most methods are relatively minor. Nevertheless, our method still achieves the best results, surpassing all baselines with a notable improvement. Furthermore, the performance curves on ImageNet-R and Cars196 in 5-, 10- and 20-task settings are depicted in \Cref{fig:curve}, illustrating that our method significantly outperforms others at almost every incremental tasks with a noticeable margin.
\begin{table}[!t]
    \centering
    \footnotesize
    \renewcommand{\arraystretch}{0.8}
    \caption{Experimental results on Mini-ImageNet100 with 5- and 10-task settings. “Memory” means whether the method requires old class samples. The subscript number represents the standard deviation.}
    \label{tab:mini-IN}
    \setlength{\tabcolsep}{3pt}
    \begin{tabular}{cccccc}
        \toprule
        \multirow{4}{*}{Method} & \multirow{4}{*}{Memory}  & \multicolumn{4}{c}{Mini-ImageNet100}  \\ 
        
        \cmidrule(l){3-6} 
        &   & \multicolumn{2}{c}{5-task} & \multicolumn{2}{c}{10-task}   \\ 
        
        \cmidrule(l){3-6} 
         &  & \textit{Last-A}  & \textit{Avg-A} & \textit{Last-A} & \textit{Avg-A}  \\ 
         
        \midrule

        PROOF & \Checkmark & 92.80\std{0.06} & 95.48\std{0.03} & 92.88\std{0.05} & 95.83\std{0.04} \\

        PR-CLIP & \Checkmark & 93.57\std{0.11} & 95.77\std{0.14} & 93.07\std{0.18} & 95.19\std{0.09} \\
        
        \cmidrule(lr){1-6}

        L2P  & \XSolidBrush  & 92.36\std{0.13}  & 95.21\std{0.06}  & 90.97\std{0.28} & 94.94\std{0.13}  \\
        
        DualPrompt  & \XSolidBrush  & 92.49\std{0.17} & 91.88\std{0.11} & 85.77\std{0.21} & 91.47\std{0.09}  \\
        
        CODA-Prompt & \XSolidBrush  & 90.05\std{0.16} & 92.02\std{0.06}  & 85.77\std{0.53} & 91.47\std{0.15}  \\
        
        EASE & \XSolidBrush  & 92.43\std{0.05}  & 94.95\std{0.01}  & 91.73\std{0.11} & 95.40\std{0.06} \\

        SLCA  & \XSolidBrush  & 92.64\std{0.36}  & 95.74\std{0.89}   & 89.13\std{1.12}   & 94.46\std{0.59}   \\

        Continual-CLIP & \XSolidBrush  & 89.99 & 92.65 & 89.99 & 93.13  \\
        
        AttriCLIP  & \XSolidBrush & 87.67\std{1.15}  & 93.05\std{0.41}  & 86.24\std{0.94} & 91.74\std{0.46}  \\
        
        MoE-Adapters  & \XSolidBrush & 92.98\std{0.13}  & 95.45\std{0.15}  & 91.92\std{0.35} & 95.24\std{0.08}  \\
        
        Lang-CL  & \XSolidBrush   & 94.07\std{0.12}  & 96.13\std{0.03}  & 93.91\std{0.18} & 96.53\std{0.09}  \\

        \cmidrule(lr){1-6}
        
        Ours  & \XSolidBrush  & \textbf{94.86\std{0.12}}  & \textbf{96.97\std{0.56}}  & \textbf{94.38\std{0.04}}  & \textbf{96.86\std{0.63}}  \\ 
        
        \bottomrule
    \end{tabular}

\end{table}

\subsection{Generalizability Study}

\begin{table*}[t]
    \centering
    \footnotesize
    \renewcommand{\arraystretch}{1}
    \caption{Experimental results based on different pre-trained weights (SigLip and OpenCLIP) on ImageNet-R and Cars196 in 10-task setting. The subscript number represents the standard deviation.}
    \label{tab:generalize}
    \setlength{\tabcolsep}{8pt}
    \begin{tabular}{cccccccccc}
        \toprule
        \multirow{2}{*}{Dataset} & & \multicolumn{4}{c}{SigLip} & \multicolumn{4}{c}{OpenCLIP}   \\

        \cmidrule(l){3-6} \cmidrule(l){7-10}
        
        \multicolumn{2}{c}{} & Zero-shot & PROOF & PR-CLIP & Ours  & Zero-shot & PROOF & PR-CLIP & Ours  \\

        \midrule
        
        

        \multirow{2}{*}{ImageNet-R}  & \textit{Last-M} & 33.18 & 42.19\std{0.22} & 42.55\std{0.56}  & \textbf{89.63\std{0.39}} & 81.56 & 82.45\std{0.38} & 82.85\std{0.17}  & \textbf{88.42\std{0.20}} \\
        
        & \textit{Avg-M}  & 38.23  & 50.56\std{0.26} & 50.58\std{0.78}  & \textbf{93.05\std{0.64}} & 86.34 &  88.26\std{0.21}  &  87.48\std{0.12}  & \textbf{92.21\std{0.18}} \\

        \cmidrule{1-10}
        
        

        \multirow{2}{*}{Cars196}  & \textit{Last-A}  & 3.99  & 59.48\std{0.48} & 49.53\std{0.55}  & \textbf{85.43\std{0.13}} & 90.23 & 91.82\std{0.19}  & 90.71\std{0.65} & \textbf{93.43\std{0.43}} \\
        
        & \textit{Avg-A}  & 6.02  & 69.92\std{0.64} & 62.68\std{0.33}  & \textbf{90.17\std{0.66}} & 94.27  & 94.78\std{0.11} & 91.43\std{0.54}  & \textbf{96.18\std{0.33}} \\
        \bottomrule
    \end{tabular}
\end{table*}

To evaluate whether the proposed method can generalize well across different pre-trained VLMs, we further conducted experiments on two kinds of popular pre-trained weights for CLIP: OpenCLIP and SigLip. OpenCLIP differs from original CLIP mainly in its use of larger, more diverse datasets like LAION for training. SigLip reduces the reliance on negative samples during training compared to original CLIP, thereby enhancing training efficiency and decreasing resource consumption. Due to the strong performance of replayed-based methods, we compared our method with PROOF and PRCLIP. As shown in~\Cref{tab:generalize}, our method significantly outperforms two approaches on SigLip. On ImageNet-R, it achieves Last-M improvements of 47.44\% and 47.08\% over PROOF and PRCLIP, respectively. Moreover, our method also shows higher results over these two methods while using OpenCLIP as pre-trained weights. These demonstrates that our method possesses a more robust generalization capability, maintaining superior performance even when transferred to different pre-trained VLMs.

\subsection{Comparison in Inference Time}

\begin{table*}[t]
    \centering
    \footnotesize
    \renewcommand{\arraystretch}{0.8}
    \caption{Inference time for a single test image on Imagenet-R in 10-task setting. “Memory” means whether the method requires old class samples. ``*'' means using multi-process parallelization.}
    \label{tab:time}
    \begin{tabular}{cccccccccccc}
        \toprule
        \multirow{2}{*}{Method} & \multirow{2}{*}{Memory}  & \multicolumn{10}{c}{Inference time (millisecond)}  \\ 
        
        \cmidrule(l){3-12} 
        & & Task 1 & Task 2 & Task 3 & Task 4 & Task 5 & Task 6 & Task 7 & Task 8 & Task 9 & Task 10 \\
         
        \midrule
        PROOF & \Checkmark & 7.36 & 6.15 & 6.57 & 6.11 & 6.10 & 6.33 & 6.47 & 6.43 & 6.50 & 6.59 \\
        
        PRCLIP & \Checkmark & 3.3 & 3.09 & 3.09 & 3.25 & 3.15 & 3.17 & 3.29 & 3.38 & 3.41 & 3.47 \\

        \cmidrule(lr){1-12}
        
        L2P & \XSolidBrush & 7.16   & 6.56   & 6.03   & 6.05   & 6.07   & 6.25   & 6.15   & 5.92   & 5.96   & 5.97   \\
        
        DualPrompt & \XSolidBrush & 7.14   & 5.97   & 6.03   & 6.36   & 6.32   & 6.05   & 6.07   & 6.53   & 6.77   & 6.34   \\
        
        CODA-Prompt & \XSolidBrush & 7.16   & 6.56   & 6.03   & 6.36   & 6.07   & 6.05   & 5.72   & 6.08   & 6.91   & 6.72   \\
        
        AttriCLIP & \XSolidBrush  & 3.76   & 5.47   & 7.34   & 9.30   & 11.84  & 13.96  & 16.05  & 18.06  & 20.11  & 22.09  \\
        
        EASE & \XSolidBrush & 8.35   & 10.74  & 12.87  & 16.04  & 18.95  & 21.78  & 24.80  & 27.64  & 31.16  & 31.71  \\

        Lang-CL & \XSolidBrush & 3.76 & 4.01 & 5.11 & 6.58 & 7.56 & 8.77 & 10.00 & 11.24 & 12.57 & 13.89 \\
        
        \cmidrule(lr){1-12}
        Ours & \XSolidBrush & 2.67   & 5.34   & 7.09   & 8.77  & 10.50  & 12.26  & 14.78  & 17.50  & 19.16  & 20.80  \\ 

        Ours* & \XSolidBrush & 3.27   & 6.67   & 6.72   & 7.27  & 7.85  & 7.79  & 8.51  & 9.44  & 9.85  & 10.17  \\ 
        
        \bottomrule
    \end{tabular}

\end{table*}

\begin{figure}[!t]
  \centering
  \includegraphics[width=0.8\columnwidth]{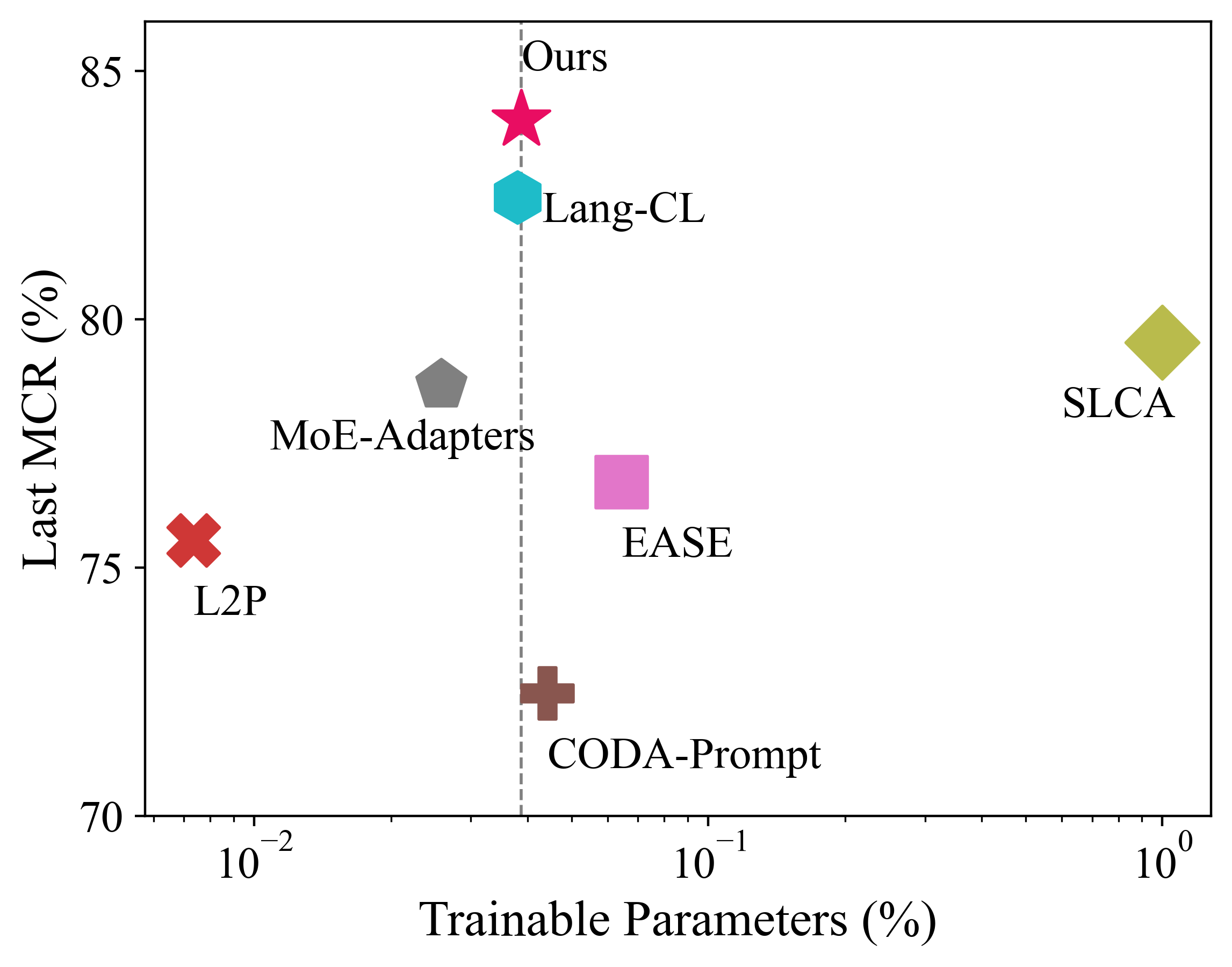}
  \caption{Trainable parameters and \textit{Last-M} on ImageNet-R of different methods.}
  \label{fig:param_ir}
\end{figure}

We evaluated the inference time for a single test sample across different methods, as summarized in~\Cref{tab:time}. While our method exhibits a slight disadvantage in inference speed, the time remains within an acceptable range. The average processing speed of 50 images per second is sufficient for most practical application scenarios that do not require high real-time performance. By trading off some efficiency, our method delivers a significant performance improvement over existing approaches. Furthermore, our model's multi-branch architecture enables simultaneous and independent execution of each branch. As a result, inference can be accelerated through multi-process parallelization, where each task-specific branch processes input images concurrently. The features extracted by all branches are then aggregated for subsequent processing. The last row of \Cref{tab:time} presents the inference time after parallelization. It is obvious that, compared to the preceding row, the inference time has significantly decreased. and as the number of tasks increases, the inference time rises at a much slower rate.

\subsection{Comparison in Trainable Parameters}

We compared the number of trainable parameters and the performance on ImageNet-R across multiple methods, as illustrated in the \Cref{fig:param_ir}. Our method achieves the best performance while maintaining a moderate number of trainable parameters. Specifically, only 11M parameters are trainable in our method, accounting for 3.9\% of the total parameter count. This demonstrates that a larger number of trainable parameters does not necessarily lead to stronger continual learning capabilities, and our method's performance advantage is not reliant on parameter quantity.

\end{document}